\documentclass[10pt,conference,a4paper]{IEEEtran}
\usepackage{ifpdf}
\usepackage{xcolor}

\ifCLASSINFOpdf
  \usepackage[pdftex]{graphicx}
  \graphicspath{{../pdf/}{../jpeg/}}
  \DeclareGraphicsExtensions{.pdf,.jpeg,.png}
\else
  \usepackage[dvips]{graphicx}
  \graphicspath{{../eps/}}
  \DeclareGraphicsExtensions{.eps}
  \fi
\usepackage{amsmath,amsxtra,amssymb,amsthm,latexsym,amscd,amsfonts}
\usepackage[utf8]{vntex}
\ifCLASSINFOpdf
\usepackage[pdftex]{graphicx}
\DeclareGraphicsExtensions{.pdf,.jpeg,.png,.jpg}
\else
\usepackage[dvips]{graphicx}
\usepackage{subfigure}

\DeclareGraphicsExtensions{.eps}
\usepackage{multirow}
\usepackage{float}
\usepackage{subcaption}

\fi
\usepackage{array}
\usepackage{epstopdf}
\usepackage{anysize}
\marginsize{2.4cm}{2.0cm}{3.2cm}{3.0cm}
\usepackage{multicol}
\usepackage{balance}
\usepackage{graphicx}
\usepackage{caption}

\makeatletter
\def\ScaleIfNeeded{\ifdim\Gin@nat@width>\linewidth\linewidth\else\Gin@nat@width\fi}

\begin{document}
\columnsep=0.63cm
\def\mathbi#1{\boldsymbol{#1}}
\def\erfc{\:\mathrm{erfc}}
\def\arg{\:\mathrm{arg}}
\def\E{\:\mathrm{E}}
\def\sinc{\:\mathrm{sinc}}
\def\T{\mathrm{T}}
\def\H{\mathrm{H}}
\newcommand{\bigsize}{\fontsize{16pt}{20pt}\selectfont}

%
\include{Abbr}
\title{Cải thiện thời gian bay dựa trên mô hình điều khiển dự đoán thích nghi cho máy bay không người lái}

\author{
\IEEEauthorblockN{
Ngô Huy Hoàng, Nguyễn Cảnh Thanh và Hoàng Văn Xiêm
} 
\IEEEauthorblockA{ Bộ môn Kỹ thuật Robot, Khoa Điện tử - Viễn Thông \\ Trường Đại học Công Nghệ - Đại học Quốc gia Hà Nội\\
		Email: ngoh52180@gmail.com, canhthanh@vnu.edu.vn, xiemhoang@vnu.edu.vn}
}
\maketitle

\begin{abstract}
Các nền tảng trên không thông minh như máy bay không người lái (UAV) đang được kỳ vọng mang đến cuộc cách mạng trong hàng loạt lĩnh vực như vận chuyển và giao thông, giám sát hiện trường, sản xuất công nghiệp, quản lý nông nghiệp. Trong đó, điều khiển chính xác là một trong những nhiệm vụ quan trọng mang tính quyết định hiệu suất và khả năng làm việc của hệ thống máy bay không người lái. Tuy nhiên, các nghiên cứu hiện nay tập trung giải quyết vấn đề theo dõi quỹ đạo, giảm thiểu sai số trong quá trình bay mà ít quan tâm tới cải thiện thời gian bay. Trong bài báo này, chúng tôi đề xuất một mô hình điều khiển dự đoán (MPC) giảm thiểu thời gian bay đồng thời khắc phục những hạn chế của bộ điều khiển MPC cổ điển thường được sử dụng. Bên cạnh đó, phương pháp MPC và ứng dụng của nó cho điều khiển máy bay không người lái đã được trình bày chi tiết trong bài báo. Cuối cùng, kết quả đã chứng minh hiệu suất của bộ điều khiển đề xuất được cải thiện so với MPC tiêu chuẩn. Ngoài ra, hướng tiếp cận này có tiềm năng trở thành nền tảng cho việc kết hợp các thuật toán thông minh vào các bộ điều khiển cơ bản.

\end{abstract}

\IEEEpeerreviewmaketitle  
\section{GIỚI THIỆU}



Máy bay không người lái (UAVs) đang tạo ra một cuộc cách mạng trong nhiều ngành công nghiệp, nông nghiệp cũng như quân sự. Nhờ khả năng di chuyển linh hoạt trong các môi trường và thiết kế nhỏ gọn, máy bay không người lái có thể cắt giảm chi phí hoạt động đi 50\%, giảm rõ rệt thời gian cần để thực hiện công việc nhất là với những hệ thống có quy mô lớn \cite{xing2023autonomous}. Để đạt được hiệu suất cao trong các ứng dụng, một yêu cầu quan trọng là phải duy trì được độ chính xác bay, trong khi giảm thời gian thực hiện chuyến bay đi ngắn nhất có thể. Điều đó đặt ra thách thức cho việc cải tiến các thuật toán điều khiển cổ điển được sử dụng, trong khi vẫn phải đảm bảo độ chính xác cũng như độ phức tạp để phù hợp cho hệ thống phần cứng hạn chế của máy bay không người lái. 

Gần đây, các phương pháp điều khiển dựa trên tối ưu hóa đặc biệt là mô hình điều khiển dự đoán (MPC) và những biến thể của nó thu hút nhiều sự chú ý cho điều khiển quadrotor nhờ những tiến bộ trong hiệu quả phần cứng, thuật toán và mô hình. MPC xem xét các giá trị hiện tại và quá khứ và tạo ra lệnh điều khiển theo đường chân trời lùi giúp tối ưu hóa sai khác trong tương lai theo đường chân trời. Các phương pháp MPC tuyến tính và phi tuyến đều đã được áp dụng để điều khiển mô hình quadrotor được tổng kết và đánh giá trong \cite{thanh2022}. Ngoài ra, MPC thể hiện khả năng hoạt động với những ràng buộc về vật lý của hệ, cũng như thích nghi tốt với hệ thống đa cảm biến, phi tuyến tính \cite{falanga2018, kamel2017, nguyen2021}. Tuy nhiên, nhiều ứng dụng của MPC vẫn gặp phải những thách thức đáng kể, chẳng hạn như yêu cầu về mô hình tính toán chính xác và sự cần thiết của việc giải những bài toán tối ưu hóa quỹ đạo trực tuyến với khả năng tính toán hạn chế của máy tính quy mô nhỏ gắn trên máy bay. 

Việc cải tiến các tham số trong bộ điều khiển MPC từ lâu đã được các nhà khoa học quan tâm đến. 
Amos trong \cite{amos2018differentiable} đề xuất MPC khả vi, khi các trọng số trong hàm chi phí được cập nhật qua từng bước thời gian để thích nghi và nâng cao hiệu suất của bộ điều khiển. Angel trong \cite{romero2022model} đã đề xuất và ứng dụng bộ điều khiển dự đoán biên mô hình, bằng sự kết hợp giữa MPC và các thành phần sai số biên \cite{lam2010model}, vừa xét tới tính chính xác của bộ điều khiển cũng như cải thiện thời gian bay. Nhìn chung, các thuật toán MPC cải tiến mặc dù đã được chứng minh là mang lại hiệu suất bay tốt hơn, nhưng vẫn có sự đánh đổi nhất là về độ phức tạp của thuật toán cải tiến, và phần lớn các thuật toán MPC cải tiến cho quadrotor chưa thể hoạt động được trong điều kiện thời gian thực \cite{foehn2021time}.

 Cải tiến thời gian tối ưu trong các thuật toán điều khiển là một hướng nghiên cứu tiềm năng trong tương lai. Các thuật toán điều khiển cơ bản mặc dù khai thác nhiều vào sai số giữa trạng thái và tín hiệu điều khiển hiện tại đối với trạng thái và tín hiệu điều khiển tham chiếu nhưng rất ít thuật toán xét tới việc tối ưu hoá thời gian thực hiện. 
 Neunert và các cộng sự trong \cite{neunert2016} đã đưa ra một mô hình MPC với một trong những đầu vào là thời gian tối ưu để điều khiển UAV. Thời gian tối ưu trong bài báo này được định nghĩa là thời gian từ khi bắt đầu chuyển động cho tới lúc đạt được một điểm tham chiếu quan trọng (thường là điểm đích).




    
\begin{figure}[!ht]
    \centering
    \includegraphics[width=0.5\textwidth]{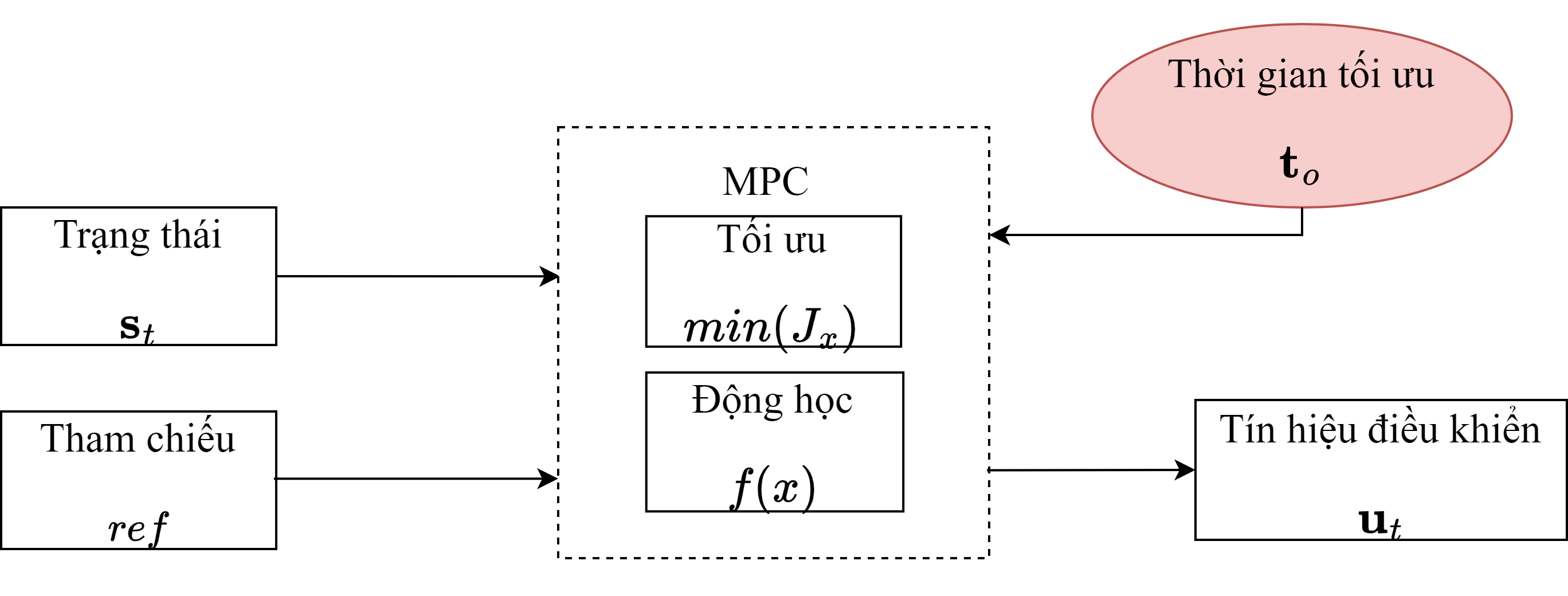}
    \caption{Tổng quan về thuật toán cải tiến đề xuất}
    \label{fig:overview}
\end{figure}

Trong bài báo này, chúng tôi đề xuất thuật toán MPC cải tiến nhằm tối ưu thời gian bay cho mô hình máy bay không người lái được mô tả trong Hình \ref{fig:overview}. Ngoài ra, chúng tôi cũng ứng dụng và trình bày toàn diện mô hình MPC trong điều khiển máy bay không ngưới lái để phù hợp với đặc trưng động học và động lực học của hệ thống. Cuối cùng, chúng tôi đánh giá thuật toán đề xuất với thuật toán MPC tiêu chuẩn, cho thấy hiệu quả trong việc đáp ứng vị trí, thời gian bay, góc lệch và đưa ra một số hướng phát triển tiềm năng của thuật toán trong tương lai. 

Cấu trúc của bài báo được sắp xếp theo thứ tự như sau: Hệ thống điều khiển đề xuất được mô tả trong phần II. Tiếp đến, phần III trình bày những kết quả đạt được, bao gồm cả việc cấu hình hệ thống với những ràng buộc động học và động lực học, cùng với đó là kết quả và đánh giá hiệu năng của thuật toán MPC cải tiến. Cuối cùng, những kết luận được nêu rõ trong phần IV.

\section{Đề xuất bộ điều khiển MPC cải tiến}
\subsection{Bộ điều khiển MPC tiêu chuẩn}
\label{Sec:LMPC}
Một mô hình MPC tiêu chuẩn thường được chia ra là hai loại: MPC tuyến tính và MPC phi tuyến. Trong khi MPC tuyến tính xấp xỉ hoá mô hình động học - động lực học và rời rạc hoá thời gian nhằm tối giản mô hình, MPC phi tuyến xét tới mô hình một cách đầy đủ và toàn diện, và thường đem lại hiệu suất bay tốt hơn so với MPC tuyến tính. Chi tiết hơn về bộ điều khiển MPC cũng như tương tác giữa không gian trạng thái của MPC và mô hình động học cuả máy bay không người lái được chúng tôi trình bày trong \cite{thanh2022}.
\subsubsection{MPC tuyến tính}
MPC tuyến tính được xây dựng như một bài toán tối ưu hoá phương trình bậc hai. Chúng tôi định nghĩa vector trạng thái $\mathbi{x} \in \mathbb{R}^{12}$ như sau:
\begin{equation}
    \mathbi{x} = \{\mathbi{\xi}^T, \mathbi{\eta}, \dot{\mathbi{\xi}}^T, \dot{\mathbi{\eta}}^T\}.
\end{equation}
và vector đầu vào $\mathbi{u} \in \mathbb{R}^{4}$:
\begin{equation}
    \mathbi{u} = \{\omega_1^2, \omega_2^2, \omega_3^2, \omega_4^2\}.
\end{equation}

Về cơ bản, MPC tuyến tính rời rạc hoá bước thời gian và không gian để giảm độ phức tạp cũng như cải thiện độ ổn định của hệ thống. Mô hình không gian trạng thái tuyến tính được xác định theo công thức:
\begin{equation}
    \mathbi{x}_{k+1} = \mathbi{A}\mathbi{x}_k + \mathbi{B}\mathbi{u}_k + \mathbi{V}_d\mathbi{F}_{e,k},
\end{equation}
trong đó $\mathbi{F}_{e,k}$ là các lực bên ngoài, $\mathbi{B}_d$ là ma trận nhiễu loạn.

Hàm chi phí mục tiêu $J(\mathbi{x}, \mathbi{u})$ được định nghĩa như sau:
\begin{equation}
\label{eq:lostfunction}
    J(\mathbi{x}, \mathbi{u}) = \sum_{k = 0}^{N-1}( \|^e\mathbi{x_k} \|^2_{\mathbi{Q}}) + \sum_{k = 0}^{N_u-1}(\|\mathbi{u}_k \|^2_{\mathbi{R}}) +\|^e\mathbi{x}_N\|^2_{\mathbi{P}},
\end{equation}

với:
\begin{equation}
\left\{\begin{matrix}
\mathbi{x}_{k+1} = f(\mathbi{x}_k, \mathbi{u}_k, \mathbi{F}_{e, k})\\ 
\mathbi{F}_{e,k+1} = \mathbi{F}_{e, k} \\
\mathbi{u}_{min} \leqslant \mathbi{u}_k \leqslant \mathbi{u}_{max} \\
\mathbi{x}_0 = \mathbi{x}(t_0), \mathbi{F}_{e, 0} = \mathbi{F}_{e}(t_0)
\end{matrix}\right. \hspace{0.2cm} \forall k \in \left[0, N-1\right],
\end{equation}

trong đó $\| \cdot \|$ biểu thị khoảng cách Euclidean. $\mathbi{Q} \geqslant 
 0$, $\mathbi{R} \geqslant 0$, $ \mathbi{P} \geqslant  0$ lần lượt là ma trận trọng số của trạng thái, đầu vào và trạng thái cuối cùng. $N$ là số bước dự đoán chân trời và $N_u$ là số bước điều khiển chân trời. $\mathbi{u}_{min}$, $\mathbi{u}_{max}$ là giới hạn dưới và giới hạn trên của tín hiệu điều khiển. 

Sai số trạng thái bay được định nghĩa bởi trạng thái hiện tại $\mathbi{x}_k$ và trạng thái tham chiếu $^r \mathbi{x}_k$:
\begin{equation}
     ^e\mathbi{x}_k = \mathbi{x}_k - ^r\mathbi{x}_k \hspace{0.5cm} \forall k 
     \in  \left[0, N\right].
\end{equation}

Trong trường hợp của chúng tôi, sáu trạng thái đầu vào $[ x, y, z, \phi, \theta, \psi]$ phải tuân theo quỹ đạo tham chiếu do số lượng tín hiệu điều khiển nhỏ hơn số lượng tham chiếu đầu ra dẫn tới không đủ bậc tự do để thực hiện bay độc lập cho tất cả đầu ra.

Phương trình (\ref{eq:lostfunction}) được viết lại như sau:
\begin{equation}
    J(\bar{\mathbi{x}}, \bar{\mathbi{u}}) = \bar{\mathbi{x}}^T \bar{\mathbi{Q}} \bar{\mathbi{x}} + \bar{\mathbi{u}}^T \bar{\mathbi{R}} \bar{\mathbi{u}} + ^e\mathbi{x}^T_N\mathbi{P} \ ^e\mathbi{x}_N
\end{equation}
trong đó, 

$\bar{\mathbi{x}} = \begin{bmatrix}
    ^e\mathbi{x}_1 & ^e\mathbi{x}_2 & \dots & ^e\mathbi{x}_N
\end{bmatrix} \in \mathbb{R}^{N}$ \\

$\bar{\mathbi{u}} = \begin{bmatrix}
    \mathbi{u}_1 & \mathbi{u}_2 & \dots & \mathbi{u}_{N_u-1}
\end{bmatrix} \in \mathbb{R}^{N_u}$ \\

$\bar{\mathbi{Q}} = \begin{bmatrix}
    Q_1 & 0 & \dots & 0 \\
    0 & Q_2 & \dots & 0 \\
    \vdots  & \vdots & \ddots & 0 \\
    0 & 0 & \dots & Q_N \\
\end{bmatrix}$ \\ \\

$\bar{\mathbi{R}} = \begin{bmatrix}
    R & 0 & \dots & 0 \\
    0 & R & \dots & 0 \\
    \vdots  & \vdots & \ddots & 0 \\
    0 & 0 & \dots & R \\
\end{bmatrix}$

\subsubsection{MPC phi tuyến}
\label{Sec:NMPC}
Bên cạnh bộ điều khiển MPC tuyến tính, bộ điều khiển MPC phi tuyến cho phép tính toán thời gian liên tục cho quadrotor dựa trên vấn đề điều khiển tối ưu.



Tương tự như bộ điều khiển MPC tuyến tính, chúng tôi xây dựng hàm chi phí phi tuyến như sau:
\begin{equation}
\min_{\mathbi{u}} \int_{t = 0}^{T} \bigg( \|^e\mathbi{x}(t) \|^2_{\mathbi{Q}} + \| \mathbi{u}(t)\|^2_{\mathbi{R}} \bigg) dt + \|  ^e\mathbi{x}(T)\|^2_{\mathbi{P}}
\label{eq:costNMPC}
\end{equation}

phụ thuộc vào:
\begin{equation}
\left\{\begin{matrix}
\dot{\mathbi{x}} = \mathbi{f}(\mathbi{x}, \mathbi{u})\\ 
\mathbi{u}_{min} \leqslant \mathbi{u}(t) \leqslant \mathbi{u}_{max} \\
\mathbi{x}_0 = \mathbi{x}(t_0), 
\end{matrix}\right.,
\end{equation}

với, $\| \cdot \|$ biểu thị khoảng cách Euclidean. $\mathbi{Q} \geqslant 
 0$, $\mathbi{R} \geqslant 0$, $ \mathbi{P} \geqslant  0$ lần lượt là ma trận trọng số của trạng thái, đầu vào và trạng thái cuối cùng. T chiều dài đường dự đoán. $\| \mathbi{u}(t)\|^2_{\mathbi{R}}$ là giá trị đầu vào,  $\| ^e\mathbi{x}(T)\|^2_{\mathbi{P}}$  đánh giá độ lệch so với trạng thái mong muốn ở cuối đường dự đoán .$\|^e\mathbi{x}(t) \|^2_{\mathbi{Q}}$ là giá trị trạng thái được định nghĩa bằng sai số của trạng thái hiện tại $\mathbi{x}(t)$ và trạng thái tham chiếu $^r\mathbi{x}(t)$:
 \begin{equation}
     ^e \mathbi{x}(t) = \mathbi{x}(t) - ^r \mathbi{x}(t)
 \end{equation}

Bộ điều khiển thực hiện tối ưu hóa theo kiểu đường chân trời lùi dần. Bên cạnh đó, các ràng buộc về tính liên tục được áp đặt vào trong hệ thống động lực học.
\subsection{Bộ điều khiển MPC cải tiến thời gian tối ưu}
Mặc dù MPC tuyến tính và phi tuyến đã đạt được hiệu suất ổn định cả trong mô phỏng cũng như thực tế, bộ điều khiển tồn tại những hạn chế lớn. MPC tập trung vào việc tối thiểu hoá sai số giữa máy bay và đích, dự đoán trước tương lai gần cũng như đảm bảo các ràng buộc về động học, động lực học và ràng buộc về động cơ; do vậy MPC chỉ có thể giải quyết tốt bài toán theo dõi quỹ đạo - khi đường bay đã được lập sẵn qua một tập hợp điểm được định trước. Tuy nhiến, đối với bài toán lập đường đi và bay từ điểm tới điểm trong không gian, MPC chưa đem lại hiệu quả vượt trội và hơn hết, thời gian thực hiện thành công những chuyến bay thường không được tối ưu và xem xét. Do vậy, chúng tôi tập trung giải quyết vấn đề thời gian tối ưu thông qua cải tiến hàm mất mát.

Chúng tôi định nghĩa vector trạng thái và vector đầu vào của MPC cải tiến tương đồng với MPC phi tuyến như sau:
\begin{equation}
    \mathbi{x} = \{\mathbi{\xi}^T, \mathbi{\eta}, \dot{\mathbi{\xi}}^T, \dot{\mathbi{\eta}}^T\}.
\label{eq:stateEq}
\end{equation}
\begin{equation}
    \mathbi{u} = \{\omega_1^2, \omega_2^2, \omega_3^2, \omega_4^2\}.
    \label{eq:inputEq}
\end{equation}
Chúng tôi cải tiến bộ MPC phi tuyến ở công thức \ref{eq:costNMPC} bằng cách thêm một thành phần vào hàm chi phí mục tiêu J. Ngoài ba thành phần dã được định nghĩa và giải thích, chúng tôi định nghĩa $J_{i}$ như sau:
\begin{equation}
J_i = \sum_{k=0}^{N-1}  \left( |e\mathbi{x}_k|^2 \right) \cdot Q_i \cdot \left| e^{-\alpha(t-t_o)} - 1 \right|
\label{impv}
\end{equation}
trong đó $\| \cdot \|$ biểu thị khoảng cách Euclidean. $\mathbi{Q} \geqslant 
 0$, $\mathbi{R} \geqslant 0$, $ \mathbi{P} \geqslant  0$ lần lượt là ma trận trọng số của trạng thái, đầu vào, trạng thái cuối cùng và thời gian tối ưu. $N$ là số bước dự đoán chân trời và $N_u$ là số bước điều khiển chân trời. $\mathbi{u}_{min}$, $\mathbi{u}_{max}$ là giới hạn dưới và giới hạn trên của tín hiệu điều khiển. $t$ biểu thị thời gian hiện tại của hệ thống, trong khi đó $t_o$ là thời gian tối ưu được định nghĩa và $\alpha$ là hằng số.
Như vậy, hàm chi phí mục tiêu mới sẽ được định nghĩa là:
\begin{equation}
J(\mathbi{x}, \mathbi{u}, \mathbi{t_o}) = J_x + J_u + J_p + J_i
\end{equation}

trong đó, $J_x$ là thành phần sai số theo trạng thái,$J_u$ là thành phần sai số theo tín hiệu điều khiển, $J_p$ là thành phần sai số theo trạng thái dự đoán, $J_i$ là thành phần sai số theo thời gian tối ưu.

 Thành phần $J_i$ có hai ý nghĩa: đầu tiên, tham số e mũ biểu thị $J_i$ là một hàm có trọng số giảm dần, và sẽ đạt cực tiểu tại thời điểm $t = t_o$. Điều này có ý nghĩa là nếu thời gian bay tổng thể thực tế càng gần $t_o$, thành phần $J_i$ sẽ càng có giá trị vì nếu $t>t_o$, $J_i$ sẽ lại bắt đầu lớn hơn, làm hàm mục tiêu trở nên lớn hơn không cần thiết và gây ra những sai lệch trong quá trình ước lượng và tối ưu hoá. Thứ hai, một thành phần giống với thành phần trạng thái $J_x$ được lựa chọn để làm số nhân với $J_i$. Thành phần giống với $J_x$ này lúc đầu sẽ có giá trị lớn do sai số ở những thời điểm ban đầu luôn lớn, tuy nhiên sẽ nhỏ dần theo thời gian. Việc thêm số nhân này kích thích cho bộ điều khiển sinh ra những lệnh điều khiển có độ lớn cao hơn trong những bước thời gian đầu của chuyến bay, và sẽ giảm dần trong quá trình bay. Điều này đặc biệt phù hợp với bài toán máy bay không người lái bay từ điểm tới điểm, khi không cần phải bay theo chuẩn xác một quỹ đạo định từ trước thì máy bay sẽ ưu tiên việc tạo ra những chuyển động nhanh trong khoảng thời gian mới bay để nhanh chóng tiếp cận hướng tới đích. 

Việc thêm một thành phần $J_i$ không làm ảnh hưởng tới những thành phần khác trong hàm chi phí hay sự hội tụ của toàn bộ hàm số. Trực quan hoá, thành phần $J_i$ có tác dụng giống việc dịch toàn bộ hàm số $J$ sang phải ở những bước thời gian đầu trong khi vẫn giữ nguyên các tính chất khác, từ đó kích thích hệ thống đưa ra những tín hiệu điều khiển cao hơn trong khoảng thời gian này. 

\subsection{Ứng dụng MPC cho điều khiển máy bay không người lái}
\begin{figure}[!ht]
    \centering
    \includegraphics[width=0.5\textwidth]{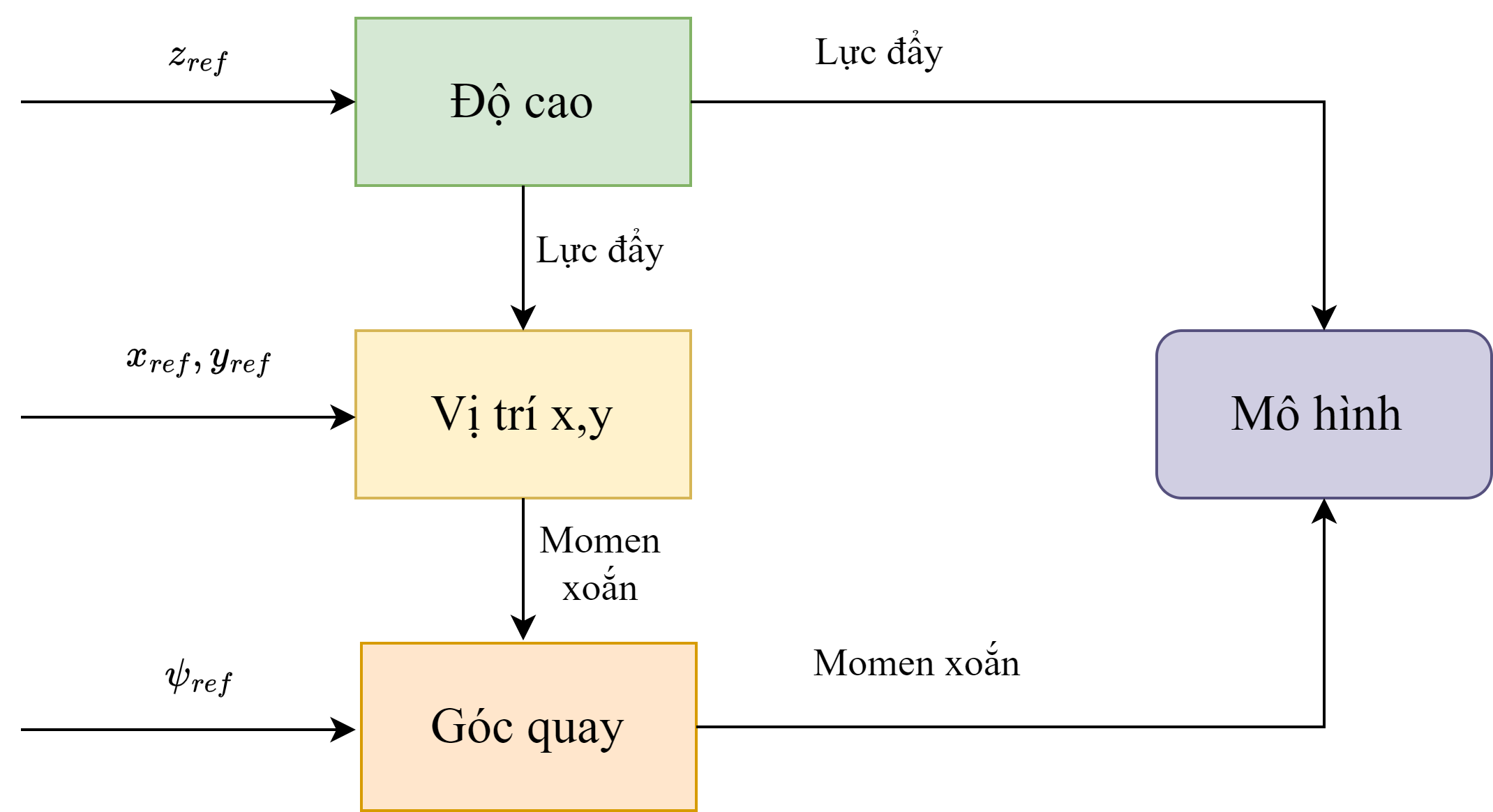}
    \caption{Bộ điều khiển MPC cho quadrotor}
    \label{fig:controller}
\end{figure}

Dựa theo những đặc trưng về phương trình động học và động lực học của máy bay không người lái, chúng tôi đề xuất một hệ thống điều khiển toàn diện cho máy bay không người lái. Theo đó trong những nghiên cứu trước về ứng dụng MPC cho điều khiển máy bay không người lái thường ứng dụng trực tiếp một bộ MPC để điều khiển cho tất cả các giá trị trạng thái và thời gian \cite{luukkonen2011modelling, thanh2022}. Chúng tôi sử dụng 3 bộ diều khiển MPC cho ba thành phần độ cao, toạ độ theo $x$, $y$ và góc quay. Lý do cho sự tách biệt của ba bộ điều khiển đến từ việc trong khi độ cao của quadrotor chỉ liên quan tới lực đẩy theo phương thẳng đứng của 4 động cơ, giá trị toạ độ của máy bay theo trục $x$ và $y$ còn phụ thuộc vào momen xoắn của các động cơ. Chi tiết về bộ điều khiển MPC cho máy bay không người lái được thể hiện trên hình \ref{fig:controller}.

\section{Kết quả}
\subsection{Thiết lập môi trường}
Chúng tôi mô hình bài toán của mình để điều khiển máy bay không người lái tiếp cận một điểm đích động. Phương trình của điểm đích được định nghĩa như sau:
\begin{align*}
        x &= 5 \ cos(\frac{2\pi}{5}t) \ \ (m)  \\ 
        y &= 5 \ sin(\frac{2\pi}{5}t)\ \  (m) \\
        z &= 0.5 \ \ \ \ \ \ \ \ \ \ \ \  (m)
\end{align*}
\begin{figure}[!ht]
    \centering
    \captionsetup{justification=centering}
    \includegraphics[width=0.35\textwidth]{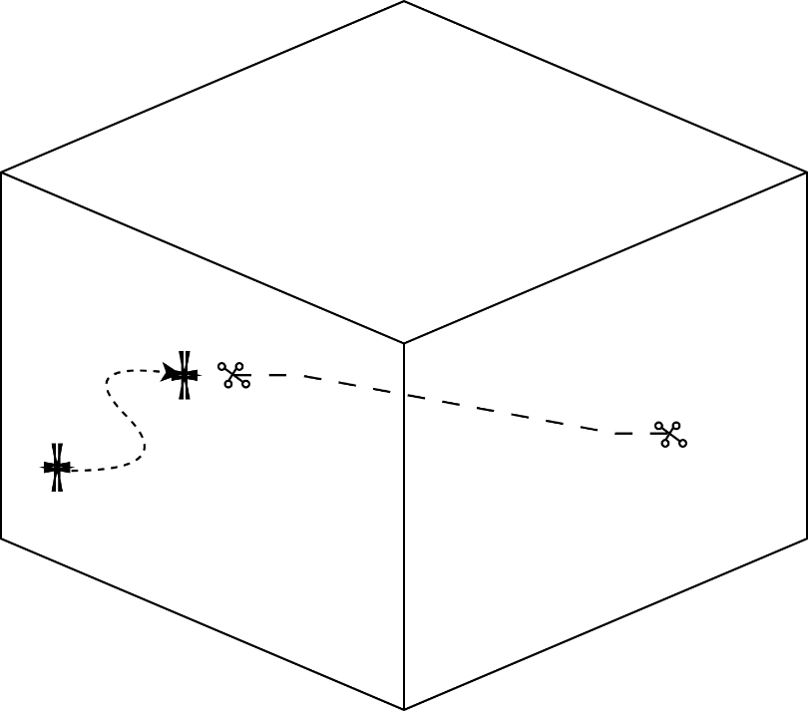}
    \caption{Mô hình bài toán}
    \label{prb_formulation}
\end{figure}
Bộ điều khiển với hàm chi phí đã được trình bày trong phần \ref{Sec:NMPC} và \ref{Sec:LMPC}. Chúng tôi giới hạn tất cả các tham số đầu vào điều khiển theo biên độ sau:
\begin{equation}
    \mathbb{U} = \left\{ \mathbi{u} \in \mathbb{R}^4 | \left[ \begin{matrix}
0 \ (rad/s)\\ 
0 \ (rad/s)\\ 
0 \ (rad/s) \\
0 \ (rad/s)
\end{matrix} \right] \leqslant \mathbi{u } \leqslant \left[ \begin{matrix}
5 \ (rad/s)\\ 
5 \ (rad/s)\\ 
5 \ (rad/s) \\
5 \ (rad/s)
\end{matrix} \right] \right\}
\end{equation}

Bên cạnh đó, tỷ lệ thay đổi của tham số đầu vào điều khiển được giới hạn nhằm ngăn chặn các chuyển động đột ngột với:
\begin{equation}
\delta\mathbi{u }_{max} = - \delta\mathbi{u }_{min}  = \left[ \begin{matrix}
1 \ (rad/s)\\ 
1 \ (rad/s)\\ 
1 \ (rad/s) \\
1 \ (rad/s)
\end{matrix} \right] 
\end{equation}
 \begin{table}[ht]
\centering
\caption{So sánh hiệu suất của MPC tiêu chuẩn và MPC cải tiến (IMPC)}
\begin{tabular}{|c|c|c|}
\hline
& MPC & IMPC \\
\hline
Tổng sai số x & 401.43 & \textbf{348.43} \\
\hline
Tổng sai số y & 404.00 & \textbf{348.43} \\
\hline
Tổng sai số z & 10.70 & \textbf{8.19} \\
\hline
Sai số x nhỏ nhất & 0.009 & \textbf{2.86e-3} \\
\hline
Sai số y nhỏ nhất & \textbf{2.64e-3} & 2.89e-3 \\
\hline
Sai số z nhỏ nhất & 4.84e-12 & \textbf{2.55e-15} \\
\hline
\end{tabular}
\label{er:all}
\end{table}

\begin{table*}[ht]
\caption{So sánh thời gian bay giữa MPC tiêu chuẩn và MPC cải tiến với các giá trị thời gian tối ưu khác nhau }
\centering
\begin{tabular}{m{0.15\textwidth} m{0.1\textwidth} m{0.1\textwidth} m{0.1\textwidth} m{0.1\textwidth} m{0.1\textwidth} m{0.1\textwidth} m{0.1\textwidth}  }
\hline
Phương pháp &LQR & MPC & $t_o = 1$ & $t_o = 2$ &$t_o = 2.4$ & $t_o = 5$& $t_o = 10$ \\ \hline

Thời gian bay(giây) &2.83& 2.75 & 2.53 & 2.45 & \textbf{2.42} & 2.49 & 2.59 \\ 
\hline
\end{tabular}
\label{time}
\end{table*}

Chúng tôi chọn ma trận trọng số như sau:
\begin{equation}
\begin{aligned}
    \mathbi{P} &= \mathbi{Q} = diag(\begin{bmatrix}
    ones(1,10) & zeros(1, 10)
    \end{bmatrix}) \\
    \mathbi{Q_i} &= diag(\begin{bmatrix}
    ones(1,15) & zeros(1, 15)
    \end{bmatrix}) \\
    \mathbi{R} &= diag(\begin{bmatrix}
    0.1 & 0.1 & 0.1 & 0.1
    \end{bmatrix})
    \end{aligned}
\end{equation}
Bên cạnh đó, số bước dự đoán (prediction horizon) và số bước điều khiển (control horizon) được đặt tương ứng với $N = 18$, $N_u = 1$.
Quadrotor được khởi tạo với vị trí ban đầu là $\mathbi{\xi} (0) = (0, 0, 0) ^T$ và vận tốc góc ban đầu là $\mathbi{\eta} (0) = (0, 0, 0) ^T$.

Cuối cùng, giá trị $\alpha$  trong công thức \ref{impv} được ấn định là 0.5.

\subsection{Kết quả mô phỏng}
Trong phần này, các mô phỏng được thực hiện để cho thấy hiệu quả của bộ điều khiển. Trong các hình vẽ, chúng tôi định nghĩa MPC là MPC tiêu chuẩn và IMPC là MPC cải tiến.

\begin{figure*}[!ht]
    \centering
    \includegraphics[width=0.31\textwidth]{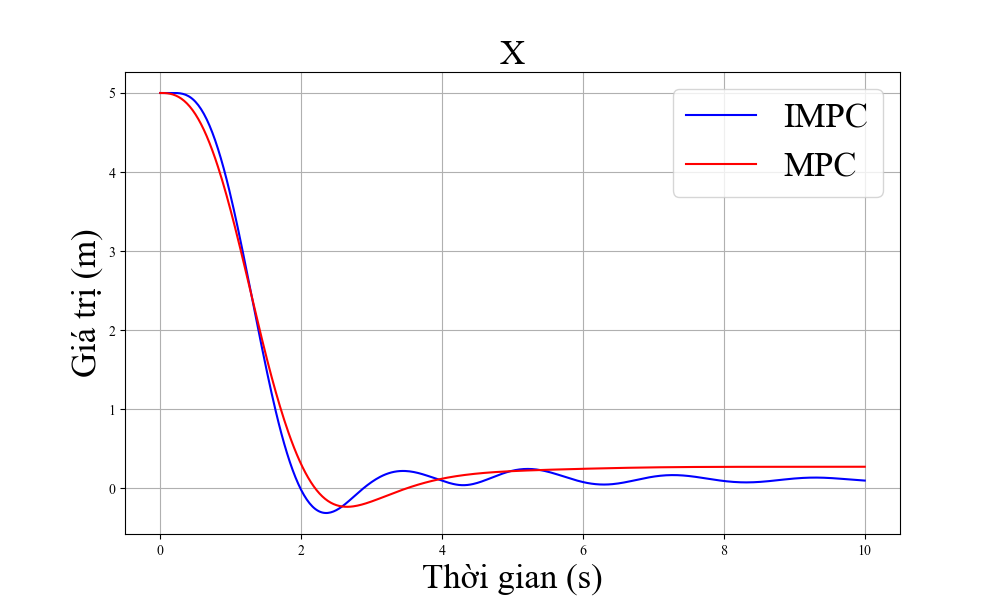}
    \includegraphics[width=0.31\textwidth]{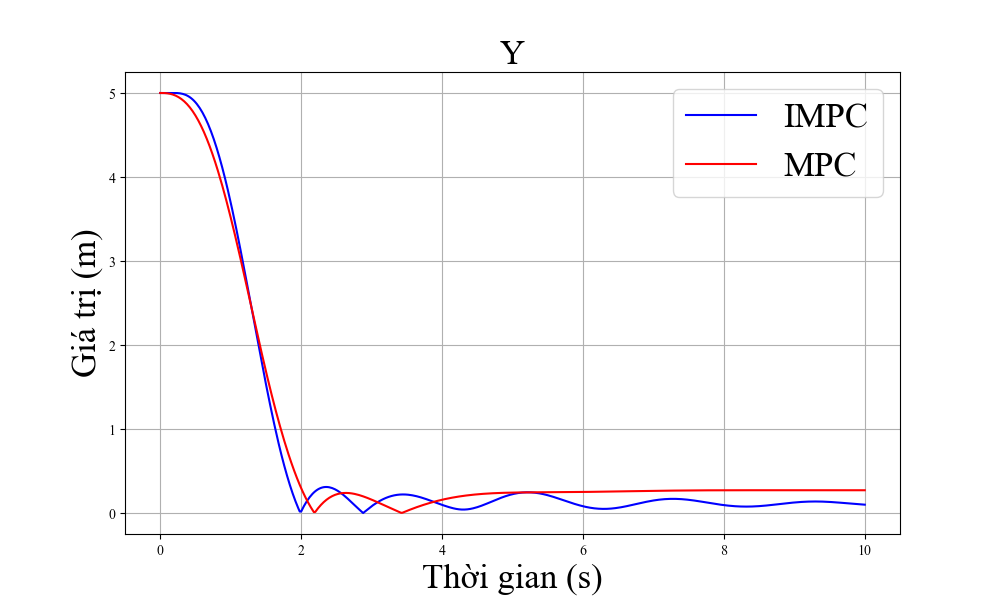}
    \includegraphics[width=0.31\textwidth]{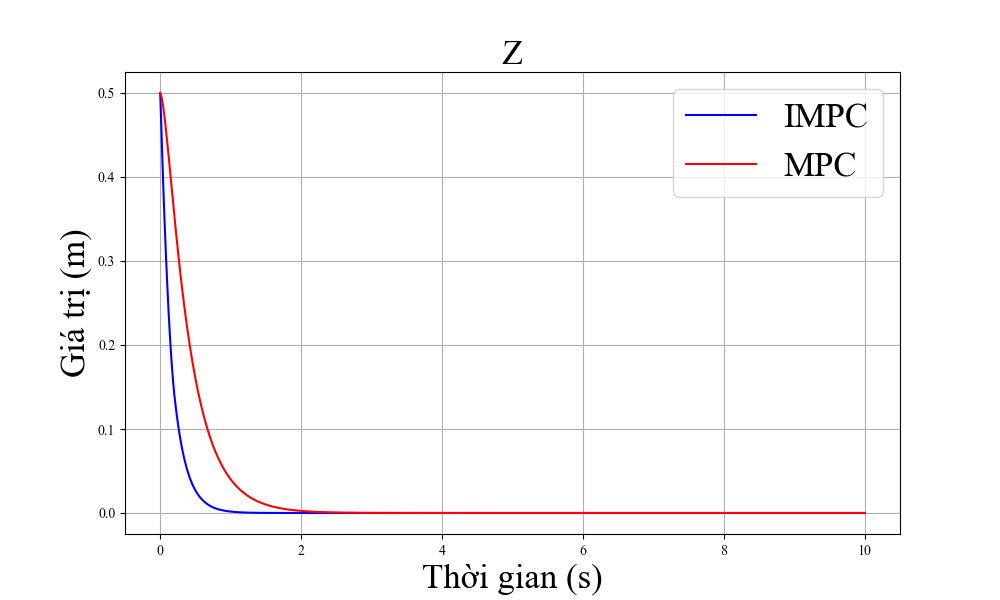}
    \includegraphics[width=0.31\textwidth]{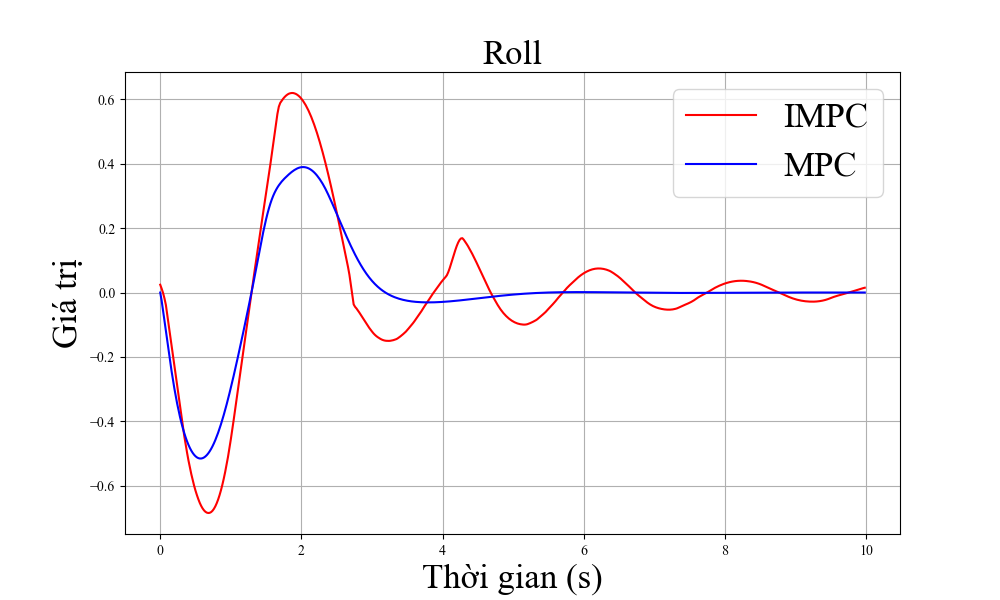}
    \includegraphics[width=0.31\textwidth]{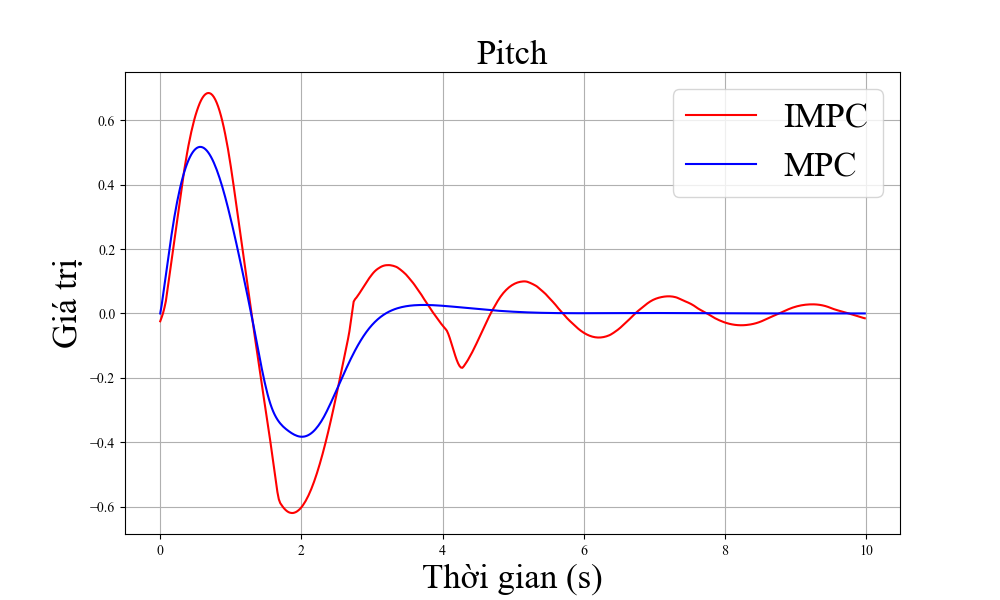}
    \includegraphics[width=0.31\textwidth]{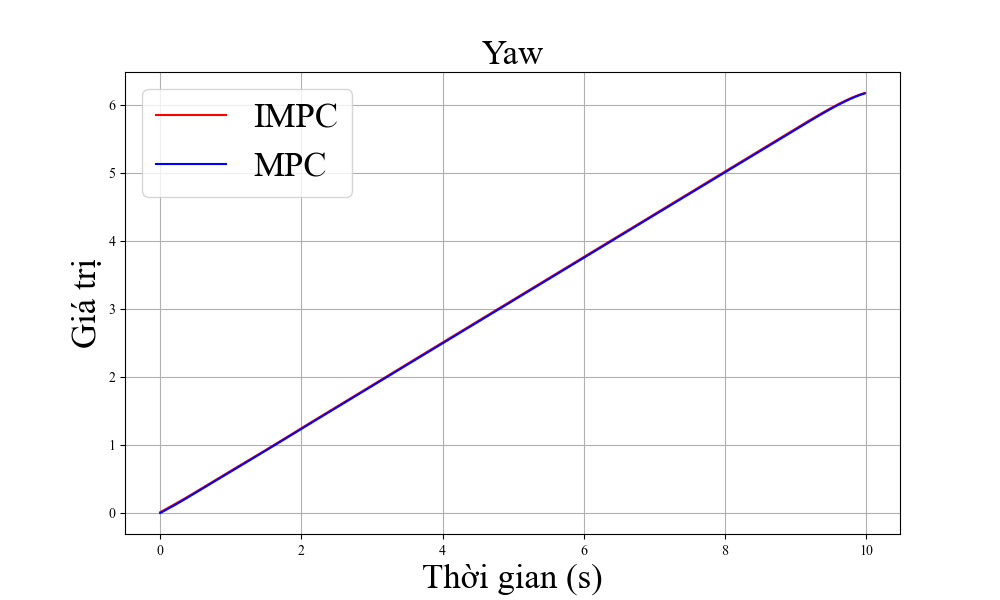}
    \caption{Sai số bám quỹ đạo theo từng thành phần}
    \label{fig:errorTrj}
\end{figure*}

Hình \ref{fig:errorTrj} cho thấy sai số của trạng thái quadrotor với trạng thái tham chiếu của hai bộ điều khiển là MPC tiêu chuẩn và MPC cải tiến. Tại thời điểm bắt đầu chuyển động, vị trí của máy bay không người lái cách xa điểm đích do đó nhiệm vụ của các bộ điều khiển là hướng tới điểm tham chiếu. 

Sai số về trạng thái của hệ thống hội tụ về 0. Bộ điều khiển MPC cải tiến cho khả năng hội tụ nhanh hơn so với bộ điều khiển MPC tiêu chuẩn, nhất là khả năng tiến tới đích theo trục z. Ngoài ra, sai số ở trạng thái ổn định của bộ điều khiển đề xuất cũng nhỏ hơn đáng kể so với bộ MPC tiêu chuẩn. Ở khả năng điều khiển góc quay, IMPC cũng cho hiệu suất tốt hơn so với MPC tiêu chuẩn nhất là ở khả năng điều khiển góc xoay roll và pitch. 

Chúng tôi định lượng những kết quả trên bằng bảng \ref{er:all}. Trong đó, tổng sai số được tính bằng tích luỹ các sai số trên các trục trong từng bước thời gian khảo sát. Sai số nhỏ nhất được tính bằng giá trị sai số nhỏ nhất trong tất cả các bước thời gian. Kết quả mô phỏng cho thấy IMPC mang lại tích luỹ sai số nhỏ hơn đáng kể trên cả ba trục x y và z, với lần lượt 14\%, 15\% và 23\%. Bên cạnh đó, sai số nhỏ nhất của phương pháp đề xuất cũng nhỏ hơn so với phương pháp cổ điển, thể hiện khả năng điều khiển chính xác được cải thiện của phương pháp MPC cải tiến so với MPC tiêu chuẩn. Tuy nhiên, sai số nhỏ nhất ở trục y của MPC tiêu chuẩn vẫn nhỏ hơn so với MPC cải tiến.

Cuối cùng, chúng tôi thử nghiệm các giá trị $t_o$ đầu vào khác nhau và biểu thị kết quả trên bảng \ref{time}. Thời gian bay ở đây được định nghĩa là thời gian ngắn nhất để máy bay từ điểm xuất phát tới điểm có sai số nhỏ hơn 0.01(m) ở cả ba trục. Nhìn chung, giá trị thời gian bay tốt nhất của phương pháp đề xuất đã cải thiện 11\% so với phương án tiêu chuẩn. Ngoài ra, nếu thời gian tối ưu được lựa chọn chuẩn xác (gần với thời gian thực tế thực hiện chuyển bay) thì hiệu suất bay cũng được cải thiện. Đây có thể là cơ sở để kết hợp các thuật toán học máy cùng với bộ điều khiển MPC cải tiến để lựa chọn tham số thời gian bay tối ưu. 

\begin{figure*}[!ht]
    \centering
    \includegraphics[width=0.4\textwidth]{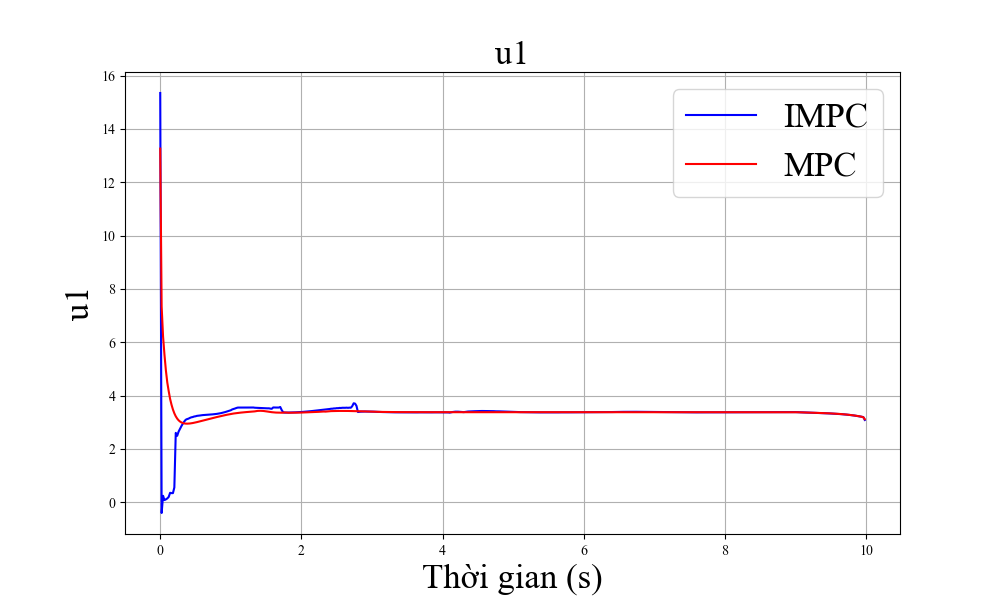}
    \includegraphics[width=0.4\textwidth]{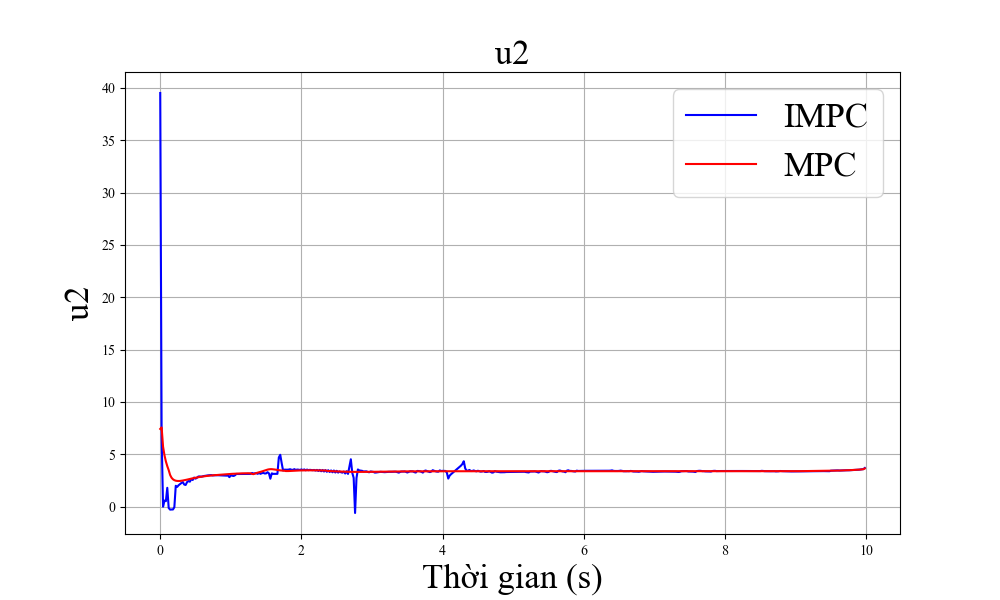}
    \includegraphics[width=0.4\textwidth]{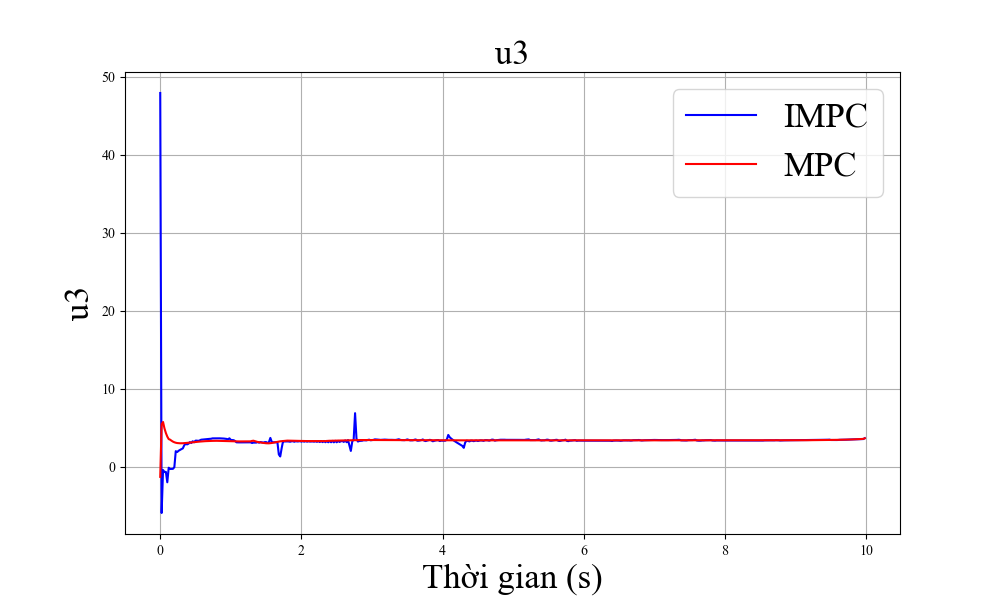}
    \includegraphics[width=0.4\textwidth]{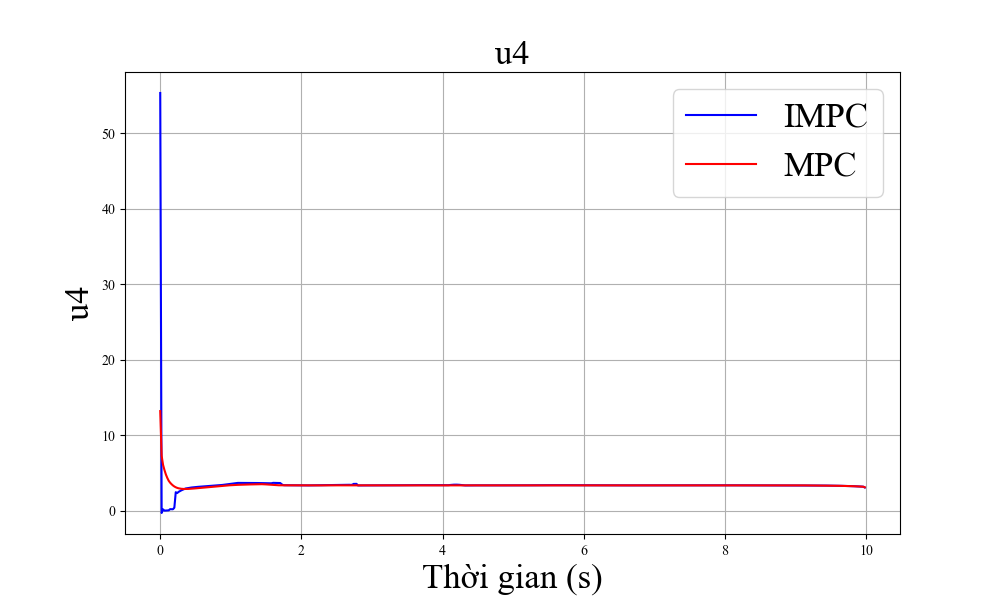}
    \caption{Tín hiệu điều khiển theo thời gian}
    \label{fig:inputState}
\end{figure*}

\section{Kết luận}
\label{Sec:KetLuan}

Trong bài báo này, chúng tôi đã trình bày một mô hình cải tiến của bộ điều khiển dự đoán mô hình để giải quyết bài toán điều khiển chính xác máy bay không người lái. Mô hình điều khiển dựa trên đặc trưng của máy bay không người lái cũng được cung cấp từ đó chứng minh tính hiệu quả của bộ điều khiển đề xuất. Kết quả mô phỏng được triển khai trên phần mềm đã so sánh, đánh giá và kết luận về hiệu suất của bộ điều khiển cải tiến đề xuất. Ngoài ra, bộ điều khiển MPC tối ưu thời gian còn cho thấy tiềm năng để trở thành nền tảng cho việc kết hợp cùng các thuật toán nâng cao.

\bibliographystyle{IEEEtran}
\balance
\bibliography{reference}
\end{document}